\documentclass[10pt,twocolumn,letterpaper]{article}

\usepackage{cvpr}
\usepackage{times}
\usepackage{epsfig}
\usepackage{graphicx}
\usepackage{amsmath}
\usepackage{amssymb}

\usepackage{amsmath}
\usepackage{graphics}
\usepackage{graphicx}
\usepackage{amssymb}
\usepackage{algorithm}
\usepackage{algorithmic}
\usepackage{mathrsfs}
\usepackage{booktabs}
\usepackage{bm}
\usepackage{bbm}
\usepackage{float}
\usepackage{url}
\usepackage{amsfonts}
\usepackage[square,sort&compress,numbers]{natbib}
\usepackage{multirow}
\usepackage{textcomp}
\usepackage{supertabular}
\usepackage{array}
\usepackage{caption}
\usepackage{gensymb}

\usepackage[breaklinks=true,bookmarks=false]{hyperref}

\cvprfinalcopy 

\def\cvprPaperID{5946} 

\ifcvprfinal\fi
\begin{document}

\urlstyle{rm}

\newcolumntype{L}[1]{>{\raggedright\arraybackslash}p{#1}}
\newcolumntype{C}[1]{>{\centering\arraybackslash}p{#1}}
\newcolumntype{R}[1]{>{\raggedleft\arraybackslash}p{#1}}

\title{Castle in the Sky: Dynamic Sky Replacement and Harmonization in Videos}

\author{
Zhengxia Zou\\
University of Michigan, Ann Arbor\\
{\tt\small zzhengxi@umich.edu}
}

\twocolumn[{%
\renewcommand\twocolumn[1][]{#1}%
\maketitle
\vspace{-1em}
\centering{\includegraphics[width=\linewidth]{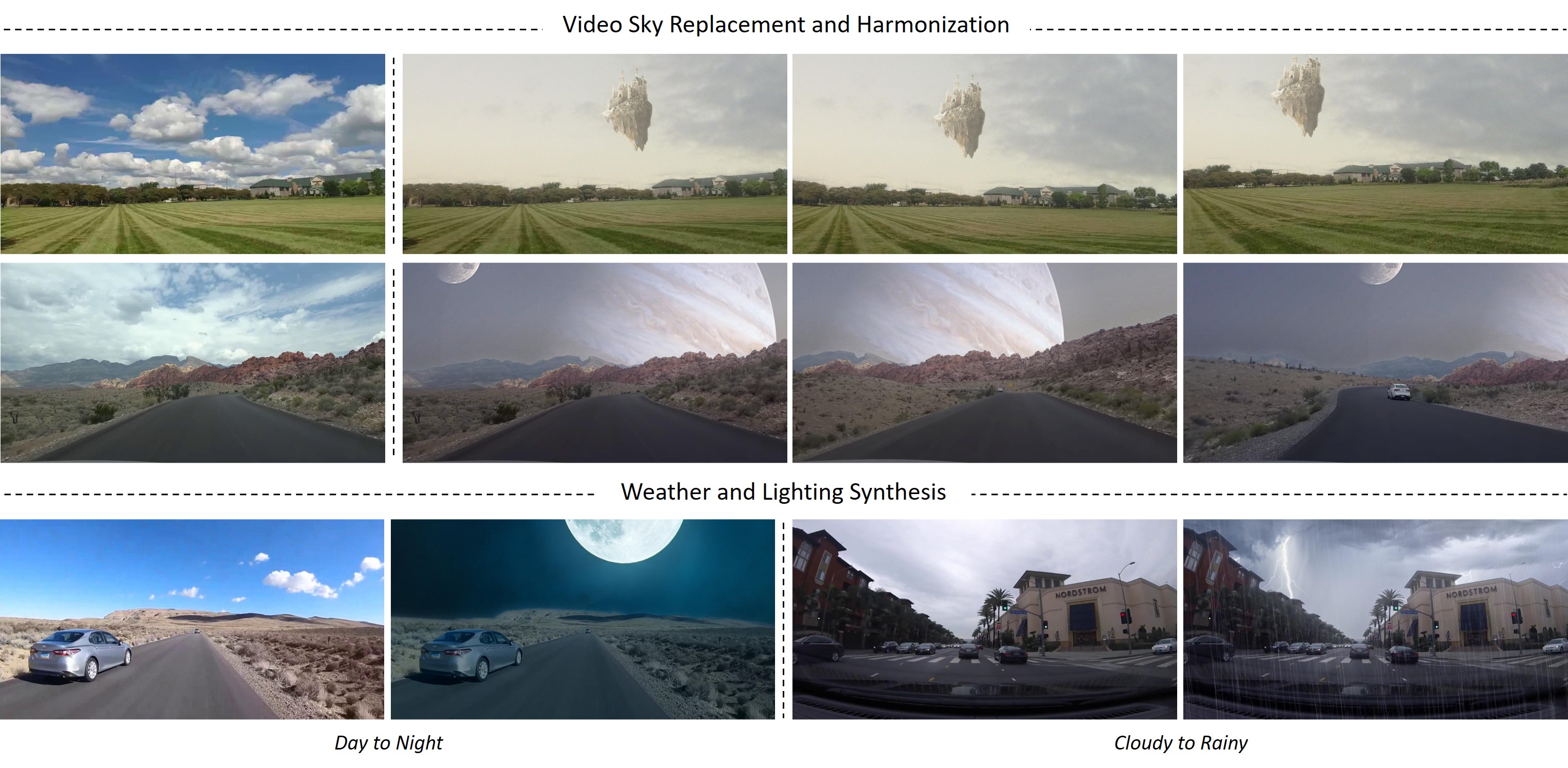}}
\captionof{figure}{\small We propose a vision-based method for generating videos with controllable sky backgrounds and realistic weather/lighting conditions. First and second row: rendered sky videos - flying castle and Jupiter sky (leftmost shows a frame from input video and the rest are the output frames). Third row: weather and lighting synthesis (day to night, cloudy to rainy).}
\label{fig:teaser}
\vspace{2em}
}]

\begin{abstract}
\vspace{-0.5em}
This paper proposes a vision-based method for video sky replacement and harmonization, which can automatically generate realistic and dramatic sky backgrounds in videos with controllable styles. Different from previous sky editing methods that either focus on static photos or require inertial measurement units integrated in smartphones on shooting videos, our method is purely vision-based, without any requirements on the capturing devices, and can be well applied to either online or offline processing scenarios. Our method runs in real-time and is free of user interactions. We decompose this artistic creation process into a couple of proxy tasks including sky matting, motion estimation, and image blending. Experiments are conducted on videos diversely captured in the wild by handheld smartphones and dash cameras, and show high fidelity and good generalization of our method in both visual quality and lighting/motion dynamics. Our code and animated results are available at \url{https://jiupinjia.github.io/skyar/}.

\end{abstract}

\section{Introduction}

The sky is a key component in outdoor photography. Photographers in the wild sometimes have to face uncontrollable weather and lighting conditions. As a result, the photos and videos captured may suffer from an over-exposed or plain-looking sky. Let's imagine if you were on a beautiful beach but the weather was bad. You took several photos and wanted to fake a perfect sunset. Thanks to the rapid development of computer vision and augmented reality, the story above is not far from reality for everyone. In the past few years, we saw automatic sky editing/optimization as an emerged research topic~\cite{tao2009skyfinder, tsai2016sky, rawat2018find, liba2020sky}. Some recent photo editors like  MeituPic\footnote{\url{https://www.meitu.com/en/}} and Photoshop\footnote{\url{https://www.youtube.com/watch?v=K_jWJ7Z-tKI}} (coming soon) have featured sky replacement toolboxes where users can easily swap out the sky in their photos with only a few clicks.

With the explosive popularity of short videos on social platforms, people are now getting used to recording their daily life with videos instead of photos. Despite the huge application needs, automatic sky editing in videos is still a rarely studied problem in computer vision. On one hand, although sky replacement has been widely applied in film production, manually replacing the sky regions in the video is laborious, time-consuming and even requires professional post-film skills. The editing typically involves frame-by-frame blue screen matting and background motion capturing. The users may spend hours on such tasks after considerable practice, even with the help of professional software. On the other hand, sky augmented reality started to be featured in recent mobile Apps. For example, the stargazing App StarWalk2\footnote{\url{https://starwalk.space/en}} can help users track stars, planets, constellations, and other celestial objects in the night sky with interactive augmented reality. Also, in a recent work of Tran \etal~\cite{Tran2020Fakeye}, a method called ``Fakeye'' is proposed for real-time sky segmentation and blending in mobile devices. However, the above approaches have critical requirements on camera hardware and cannot be directly applied to offline video processing tasks. Typically, to calibrate the virtual cameras to the real-world, these approaches require real-time pose signals of the camera from the Inertial Measurement Units (IMU) integrated with the camera like the gyroscope in smartphones.

In this paper, we investigate an interesting question that whether the sky augmentation in videos can be realized in a purely vision-based manner, and propose a new solution for such a task. As we mentioned above, previous methods of this research topic either focus on static photos~\cite{tao2009skyfinder,tsai2016sky,rawat2018find} or require the gyroscope signals available along with the video frames captured on-the-fly~\cite{Tran2020Fakeye}. Our method, different from the above ones, has no specifications on capturing devices and is suitable for both online augmented reality and offline video editing applications. The processing is ``one click and go'' and no user interactions are needed. 

Our method consists of multiple components:

- \textbf{A sky matting network} for detecting sky regions in video frames. Different from previous methods that frame this process as a binary pixel-wise classification (foreground v.s. sky) problem, we design a deep learning based coarse-to-fine prediction pipeline that produces soft sky matte for a more accurate detection result and a more visually pleasing blending effect.

- \textbf{A motion estimator} for recovering the motion of the sky. The sky video captured by the virtual camera needs to be rendered and synchronized under the motion of the real camera. We suppose the sky and the in-sky objects (e.g., sun, clouds) are located at infinity and their movement relative to the foreground is Affine. 

- \textbf{A skybox} for sky image warping and blending. Given a foreground frame, a predicted sky matte, and the motion parameters, the skybox aims to warp the sky background based on the motion and blend it with the foreground. The skybox also applies relighting and recoloring to make the blending result more visually realistic in its color and dynamic range.

We test our method on outdoor videos diverse captured by both dash cameras and handheld smartphones. The results show our method can generate high fidelity and visually dramatics sky videos with very good lighting/motion dynamics in outdoor environments. Also, by using the proposed sky augmentation framework we can easily synthesize different weather and lighting conditions.

The contribution of our paper is summarized as follows:
\begin{itemize}
\item We propose a new framework for sky augmentation in outdoor videos. Previous methods on this topic simply focus on sky augmentation/edit on static images and rarely consider dynamic videos.
\item Different from previous methods on outdoor augmented reality which require real-time camera pose signal from IMU, we propose a purely vision-based solution on such a task and applies to both online and offline application scenarios.
\end{itemize}

\begin{figure*}
    \centering{\includegraphics[width=\linewidth]{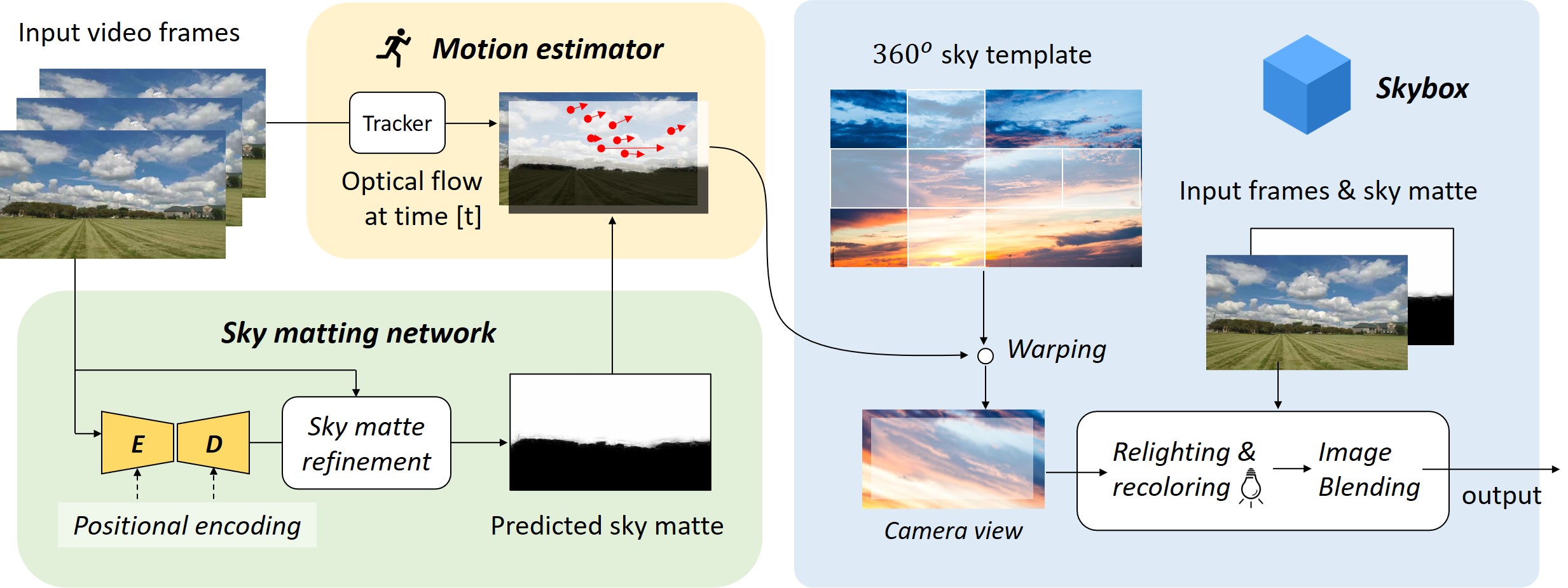}}\\
    \caption{An overview of our method. Our method consists of a sky matting network for sky matte prediction, a motion estimator for background motion estimation, and a skybox for blending the user-specified sky template into the video frames.}
    \label{fig:overview}
\end{figure*}

\section{Related Work}

Sky replacement and editing are common in photo processing and film post-production. In computer vision, some fully automatic methods for sky detection and replacement in static photos have been proposed in recent years~\cite{tao2009skyfinder,tsai2016sky,rawat2018find}. Using these methods, users can easily customize the style of the sky with their preference, with no need to have professional image editing skills.

SkyFinder~\cite{tao2009skyfinder} proposed by Tao \etal is to our best knowledge one of the first works that involve sky replacement. This method first trains a random forest based sky pixel detector and then applies graph cut segmentation~\cite{boykov2001interactive} to produce refined binary sky masks. With the sky masks obtained, the authors further achieve attribute-based sky image retrieval and controllable sky replacement. Different from the SkyFinder in which the sky replacement is considered as a by-product, Tsai \etal~\cite{tsai2016sky} focus on solving this problem and at the same time producing high-quality sky masks by using deep CNNs and conditional random field. Apart from the sky segmentation, Rawat \etal~\cite{rawat2018find} studied the matching problem between the foreground query and background images using handcrafted color features and deep CNN features, and they assumed the sky masks are available. Mihail \etal~\cite{mihail2016sky} contribute a dataset for sky segmentation which consists of 100k images captured by 53 stationary surveillance cameras. This dataset provides a benchmark for subsequent methods on sky segmentation task~\cite{la2019segmenting,Tran2020Fakeye}. They also tested several sky segmentation methods at their time~\cite{hoiem2005geometric,lu2014two,tighe2010superparsing}, either directly or as a by-product. Very recently, Liba \etal~\cite{liba2020sky} propose a method for sky optimization in low-light photography, which also involves sky segmentation. As the authors focus on mobile image processing, they designed a UNet-like CNN architecture~\cite{ronneberger2015u} for sky segmentation and followed the method Morphnet~\cite{gordon2018morphnet} for model compression and acceleration. They also introduces a refined sky image dataset on the basis of the subset of the AED20k dataset~\cite{zhou2017scene}, which contains 10,072 images diversely captured in the wild with their high-quality soft sky mattes.

As for video sky enhancement, Tran \etal proposed a method named Fakeye very recently which rendered virtual sky in real-time on smartphones~\cite{Tran2020Fakeye}. However, their method requires the pose signal to be independently read from IMU for camera alignment, and thus cannot directly be applied to offline video files. In addition, in their method, the authors choose a naive machine learning method - the pixel-wise linear classifier on pixel colors and coordinates, which fails in a complex outdoor lighting environment. ``Clear Skies Ahead'' proposed by Halperin \etal~\cite{halperin2019clear} is another recent method that focuses on video sky augmentation. Note that although this method is also a purely vision-based approach that is somewhat similar to ours, our method differs from theirs in two aspects. First, we consider the sky pixel segmentation as a soft image matting problem which produces a more detailed and natural blending results. As a comparison, the method by Halperin \etal~\cite{halperin2019clear} simply considers the sky segmentation as a pixel-wise binary classification problem which may cause blending artifacts at the segmentation junctions. Second, Halperin \etal perform camera calibration based on tracked points between frames and further assumes the motion undergoes by the camera is purely rotational, and thus a substantial camera translation may result in inaccurate rotation estimation. In our method, instead of estimating the motion of the camera, we choose to directly estimate the motion of the sky background at the infinity and can deal with both rotation and translation of the camera very well.

\section{Methodology}

Fig.~\ref{fig:overview} shows an overview of our method. Our method consists of three modules, a sky matting network, a motion estimator, and a skybox. Below we give a detailed introduction to each of them.

\subsection{Sky Matting}

Image matting~\cite{zongker1999environment,levin2008closed,xu2017deep} is a large class of methods that aims to separate the foreground of interests from an image, which plays an important role in image and video editing. Image matting usually involves predicting a soft ``matte'' which is used to extract the foreground finely from the image, which naturally corresponds to our sky region detection process. Early works of image matting can be traced back to 20 years ago~\cite{smith1996blue,zongker1999environment}. Traditional image matting methods include sampling-based methods~\cite{gastal2010shared, he2011global, shahrian2013improving}, and propagation-based methods~\cite{levin2008closed, zheng2009learning, chen2013knn}. Recently, deep learning techniques have greatly promoted the progress of image matting researches~\cite{xu2017deep, chen2018tom}.

In our work, we take advantage of the deep Convolutional Neural Network (CNN) and frame the prediction of sky soft matte under a pixel-wise regression framework which produces both coarse and fine scale sky mattes. Our sky matting networks consist of a segmentation encoder $E$, a mask prediction decoder $D$, and a soft refinement module. The encoder aims to learn intermediate feature representations of a down-sampled input image. The decoder is trained to predict a coarse sky matte. The refinement module takes in both of the coarse sky matte and the high-resolution input and produces a refined sky matte. We use a ResNet-like~\cite{he2016deep} convolutional architecture as the backbone of our encoder (other choices like VGG~\cite{simonyan2014very} also applicable) and build our decoder as an upsampling network with several convolutional layers. Note that since the sky region usually appears at the upper part of the image, we replace the conventional convolution layers with coordinate convolution layers~\cite{liu2018intriguing} at the encoder's input layer and all the decoder layers, where the normalized y-coordinate values are encoded. We show such a simple modification brings noticeable accuracy improvement on the sky matte.

Suppose $I$ and $I_l$ represent an input image with full resolution and its down-sampled substitute. Our coarse segmentation network $f=\{E, D\}$ takes in the $I_l$ and is trained to produce a sky matte with the same resolution as $I_l$. Suppose $A_l=f(I_l)$ and $\hat{A}_l$ represent the output of $f$ and the groundtruth sky alpha matte. We train the $f$ to minimize the distance between $A_l$ and the groundtruth $\hat{A}_l$ in their raw pixel space. The objective function is defined as follows:
\begin{equation}\label{eq:F_loss}
    \mathcal{L}_f (I_l) = E_{I_l\in \mathcal{D}_l}\{ \frac{1}{N_l}\|f(I_l) - \hat{A}_l\|_2^2 \},
\end{equation}
where $\|\cdot\|_2^2$ represents the pixel-wise $l_2$ distance between two images, $N_l$ is the number of pixels in the output image, and $\mathcal{D}_l$ is a dataset consists of down-sampled images on which the model $f$ is trained.

After we obtain the coarse sky matte, we up-sample the sky matte to the original input resolution in the refinement stage. We leverage the guided filtering technique~\cite{he2012guided} to recover the detailed structures of the sky matte by referring to the original input image. The guided filter is a well-known edge/structure preserving filtering method which has better behaviors near edges and better computational complexity than other popular filters like bilateral filter~\cite{tomasi1998bilateral}. Although recent image matting literature~\cite{xu2017deep, lutz2018alphagan} shows that using upsampling convolutional architecture and adversarial training may also producing fine-grained matting output, we choose guided filter mainly because of its high efficiency and simplicity. In our method, we use the full resolution image $I$ as the Guidance image and remove its red and green channels for better color contrast. With such a configuration, the filtering transfers the structures of the guidance image to the low resolution sky matte, and produces a more detailed and a sharper result than the CNN's output with minor computational overhead. We denote $A$ as the predicted full-resolution sky alpha matte after the refinement, and $A$ can be represented as follows:
\begin{equation}
    A = f_{gf}(h(A_l), I, r, \epsilon),
\end{equation}
where $f_{gf}$ and $h$ are the guided filtering and bilinear upsampling operations. $r$ and $\epsilon$ are the predefined radius and regularization coefficient of the guided filter.


\subsection{Motion Estimation}

Instead of estimating the motion of the camera as suggested by the previous method~\cite{halperin2019clear}, we directly estimate the motion of the object at infinity and then render the virtual sky background by warping a 360\degree \ skybox template image onto the perspective window. We build a skybox for image blending. The middle-right part of Fig.~\ref{fig:overview} briefly illustrates how we blend the frame and sky template based on the estimated motion and the predicted sky matte. In our method, we assume the motion of the sky patterns is modeled by an Affine matrix $\mathbf{M}\in\mathbb{R}^{3\times 3}$. Since the objects in the sky (e.g., the clouds, the sun, or the moon) are supposed to be located at infinity, we assume their perspective transform parameters are in fixed values and are already contained in the skybox background image. We compute optical flow using the iterative Lucas-Kanade method~\cite{baker2004lucas} with pyramids, and thus a set of sparse feature points can be frame-by-frame tracked. For each pair of adjacent frames, given two sets of 2D feature points, we use the RANSAC-based robust Affine estimation to compute the optimal 2D transformation with four degrees of freedom (limited to translation, rotation, and uniform scaling). Since we focus on the motion of the sky background, we only use feature points located within the sky area to compute the Affine parameters. When there are not enough feature points detected, we run depth estimation~\cite{godard2019digging} on the current frame, and then use the feature points in the second far regions to compute the Affine parameters.

Since the feature points in the sky area usually have lower quality than those in the close-range area, we find that it is sometimes difficult to obtain stable motion estimation results solely based on setting a low tolerance on the RANSAC re-projection error. We, therefore, assume that the background motion between two frames is dominated by translation and rotation and apply kernel density estimation on the Euclid distance between the paired points in each adjacent frames. The matched points with small distance probability $P(d)<\eta$ will be removed. 

After obtaining the movement of each adjacent frame, the Affine matrix $\widetilde{\mathbf{M}}^{(t)}$ across between the initial frame and the $t$-th frame in the video can be written as the following matrix multiplication form: 
\begin{equation}
    \widetilde{\mathbf{M}}^{(t)} = \mathbf{M}^{(c)} \cdot(\mathbf{M}^{(t)} \cdot \mathbf{M}^{(t-1)} \dots \mathbf{M}^{(1)}),
\end{equation}
where $\mathbf{M}^{(i)}$ ($i=1,\dots t$) represents the estimated motion parameters between frame $i-1$ and $i$. $\mathbf{M}^{(c)}$ are the center crop parameters of the skybox template, i.e., shift and scale, depending on the field of view set by the virtual camera. The final sky background $B^{(t)}$ within the camera's perspective field at the frame $t$ can be thus obtained by warping the background template $B$ using the Affine parameters $M_t$. In our method, we use a simple way to build the 360\degree \ sky background image where we tile the image when the perspective window goes out of the image border during the warping. We set the final center cropping region of the warped sky template to the field of view of the virtual camera. We set the size of the virtual camera's view as 1/2 height x 1/2 width of the template sky image. Note that other center cropping sizes are also applicable, and using a smaller range of cropping will produce a distant view effect captured by a telephoto camera.

\subsection{Sky Image Blending}

Suppose $I^{(t)}$, $A^{(t)}$, and $B^{(t)}$ are the video frame, the predicted sky alpha matte, and the aligned sky template image at time $t$. In $A^{(t)}$, a higher output pixel value means a higher probability the pixel belongs to the sky background. Based on the image matting equation~\cite{smith1996blue}, we represent the newly composed frame $Y^{(t)}$ as the linear combination of the  $I^{(t)}$ and the background $B^{(t)}$, with $A^{(t)}$ as their pixel-wise combination weights:
\begin{equation}\label{eq:matting}
    Y^{(t)} = (1-A^{(t)})I^{(t)} + A^{(t)}B^{(t)}.
\end{equation}
Note that since the foreground $I^{(t)}$ and the background $B^{(t)}$ may have different color tone and intensity, directly performing the above combination may result in an unrealistic result. We thus apply the recoloring and relighting techniques to transfer the colors and intensity from the background to the foreground. For each video frame $I^{(t)}$, we apply the following steps make corrections before the linear combination (\ref{eq:matting}):
\begin{subequations}
\begin{align}
    \widehat{I}^{(t)} &\longleftarrow I^{(t)}+\alpha(\mu_{B(A=1)}^{(t)}-\mu_{I(A=0)}^{(t)}),\label{eq:recoloring:a}\\
    I^{(t)} &\longleftarrow \beta(\widehat{I}^{(t)}+\mu_I^{(t)}-\widehat{\mu}_I^{(t)}),\label{eq:recoloring:b}
\end{align}
\end{subequations}
where $\mu_I^{(t)}$, $\widehat{\mu}_I^{(t)}$, are the mean color pixel  value of the image $I^{(t)}$ and image $\widehat{I}^{(t)}$. $\mu_{I(A=0)}^{(t)}$ and $\mu_{B(A=1)}^{(t)}$ are the mean color pixel value of the image $I^{(t)}$ at the pixel location of sky regions and the mean color pixel value of the image $B^{(t)}$ at the pixel location of non-sky regions. $\alpha$ and $\beta$ are predefined recoloring and relighting factors. In the above correction steps, the (\ref{eq:recoloring:a}) transfers the regional colortone from the background to the foreground image while the (\ref{eq:recoloring:b}) corrects the pixel intensity range of the re-colored foreground and make it compatible with the skybox background.

\begin{figure*}
    \centering{\includegraphics[width=\linewidth]{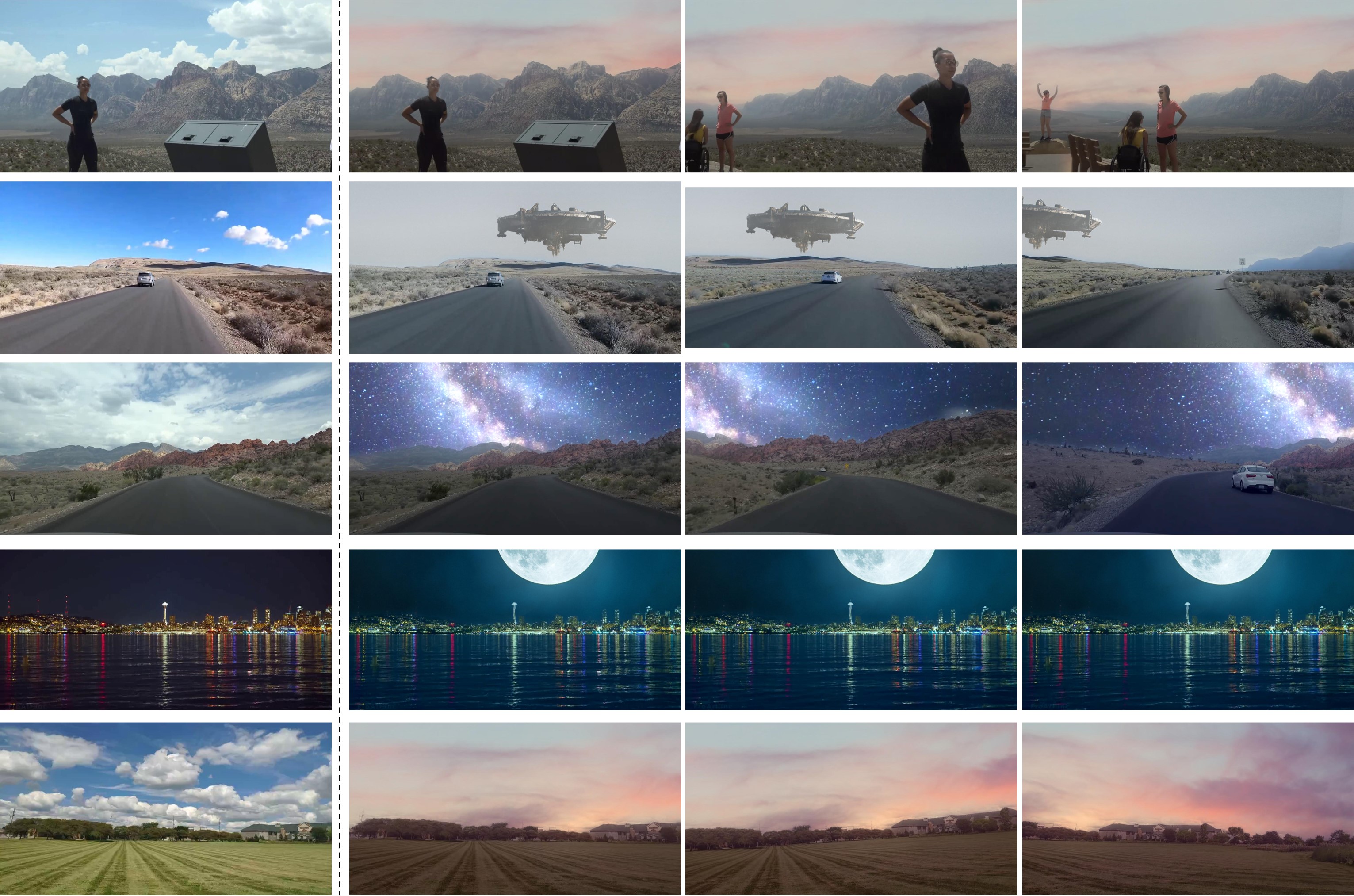}}\\
    \caption{Video sky augmentation results by using our method. Each row corresponds to a separate video clip, where the leftmost is the starting frame of the input video, and the image sequences on the right are the processed output at different time steps. We recommend readers check out our \href{https://jiupinjia.github.io/skyar/}{website} for our animated demonstration.}
    \label{fig:skyrst}
\end{figure*}

\begin{table}\small
\centering
\caption{A detailed configuration of our sky matting network. ``CoordConv'' denotes a Coordinate convolution followed by a ReLU layer. ``BN'', ``UP'', and ``Pool'' denote batch normalization, bilinear upsampling, and max pooling, respectively.}
\begin{tabular}{c|ccc}
    \toprule
     \textbf{Layer} & \textbf{Config} & \textbf{Filters} & \textbf{Reso}\\
    \midrule
     E\_1 & CoordConv-BN-Pool   & 64~/~2 & 1/4\\
     E\_2 & 3 $\times$ ResBlock & 256~/~1 & 1/4\\
     E\_3 & 4 $\times$ ResBlock & 512~/~2 & 1/8\\
     E\_4 & 6 $\times$ ResBlock & 1024~/~2 & 1/16\\
     E\_5 & 3 $\times$ ResBlock & 2048~/~2 & 1/32\\
     \midrule
     D\_1 & CoordConv-UP + E\_5  & 2048~/~1 & 1/16\\
     D\_2 & CoordConv-UP + E\_4  & 1024~/~1 & 1/8\\
     D\_3 & CoordConv-UP + E\_3  & 512~/~1 & 1/4\\
     D\_4 & CoordConv-UP + E\_2  & 256~/~1 & 1/2\\
     D\_5 & CoordConv-UP   & 64~/~1 & 1/1\\
     D\_6 & CoordConv   & 64~/~1 & 1/1\\
    \bottomrule
\end{tabular}
    \label{tab:encoderdecoder}
\end{table}

\subsection{Implementation Details}

{\bf Network architecture.} We use the ResNet-50~\cite{he2016deep} as the encoder of our sky matting networks (fully connected layers removed). The decoder part consists of five convolutional upsampling layers (coordinate conv + relu + bilinear upsampling) and a pixel-wise prediction layer (coordinate + sigmoid). We follow the configuration of the ``UNet'' \cite{ronneberger2015u} and add skip connections between the encoder and decoder layers with the same spatial size. Table~\ref{tab:encoderdecoder} shows a detailed configuration of the networks.

{\bf Dataset.} We train our sky matting network on the dataset introduced by Liba \etal~\cite{liba2020sky}. This dataset is built based on the AED20K dataset~\cite{zhou2017scene} and includes several subsets where each of them corresponds to using different methods for creating their ground truth sky matte. We use the subset ``ADE20K+DE+GF'' for training and evaluation. There are 9,187 images in the training set and 885 images in the validation set. 

{\bf Training details.} We train our model by using Adam optimizer~\cite{kingma2014adam}. We set batch size to 8, learning rate to 0.0001, and betas to (0.9, 0.999). We stop training after 200 epochs. We reduce the learning rate to its 1/10 every 50 epochs. The input image size for training is 384$\times$384. Our matting network is implemented based on PyTorch. Image augmentations we used in training include horizontal-flip, random-crop with scale=(0.5, 1.0) and ratio=(0.9, 1.1), random-brightness with brightness\_factor=(0.5, 1.5), random-gamma with $\gamma$=(0.5, 1.5), and random-saturation with saturation\_factor=(0.5, 1.5).

{\bf Other details.} In the sky matte refinement stage, we set the radius and regularization coefficient of the guided filter to $r=20$ and $\epsilon=0.01$. In the motion estimation stage, when estimating the probability of the keypoints' moving distance, we set the kernel type as ``Gaussian'' and set its bandwidth to 0.5. We set $\eta=0.1$, i.e., remove those points whose probability is smaller than 0.1.

\begin{figure*}
    \centering{\includegraphics[width=\linewidth]{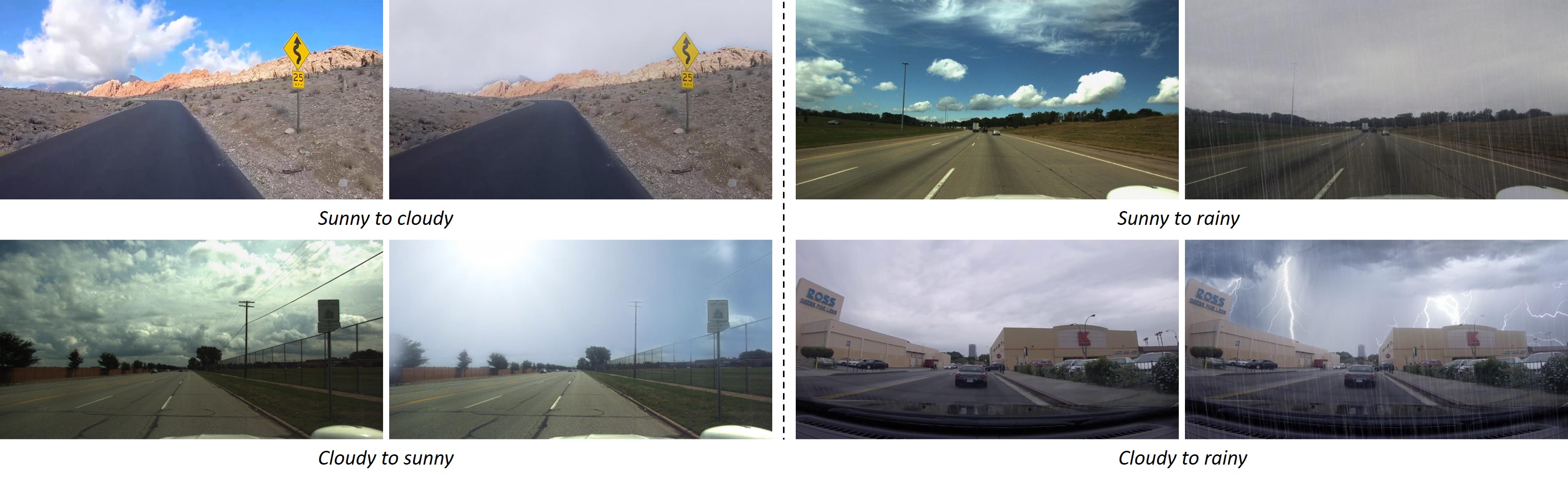}} \\
    \caption{Weather translation results by using our method. First row: sunny to cloudy, sunny to rainy. Second row: cloudy to sunny, cloudy to rainy. In each group, the left shows the original frames, and the right shows the translation result.}
    \label{fig:weatherrst}
\end{figure*}

\section{Experimental Analysis}

We evaluate our methods on video sequences diversely captured in the wild, including those by handheld smartphones and in-vehicle dash-cameras. We also test on street images from the BDD100k dataset~\cite{bdd100k}.

\begin{figure}
    \centering{\includegraphics[width=\linewidth]{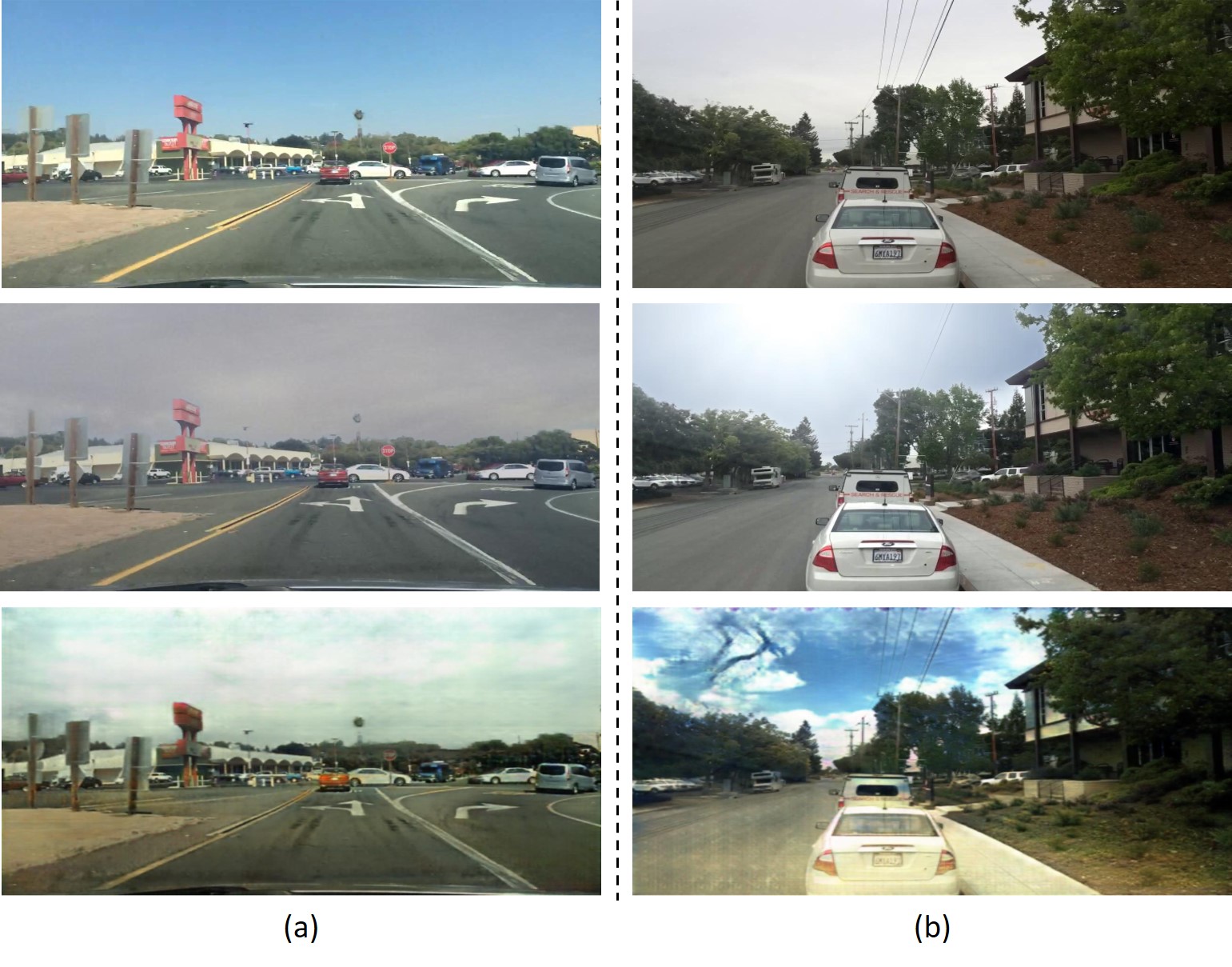}} \\
    \caption{Qualitative comparison between our method and CycleGAN~\cite{CycleGAN2017} on BDD100K~\cite{bdd100k}.  (a) Sunny to cloudy translation, (b) Cloudy to sunny translation. The first row shows two input frames. The second row shows our results. The third row shows  CycleGAN's results.}
    \label{fig:cyclegancompare}
\end{figure}

\begin{table}\small
\begin{center}
\begin{tabular}{l|cc}
\toprule
Testing scenario  & PI~\cite{blau20182018} & NIQE~\cite{mittal2012making} \\ 
\midrule
CycleGAN~\cite{CycleGAN2017} (sunny2cloudy) & 7.094 & 7.751 \\
Ours (sunny2cloudy) & 5.926 & 6.948 \\
\midrule
CycleGAN~\cite{CycleGAN2017} (cloudy2sunny) & 6.684 & 7.070 \\
Ours (cloudy2sunny) & 5.702 & 7.014 \\
\bottomrule
\end{tabular}%
\end{center}
\vspace{-1.5em}
\caption{Quantitative evaluation on the image fidelity of our method and CycleGAN under different weather translation scenarios. For both PI and NIQE, lower scores indicate better.}
\label{tab:cyclegancompare}%
\end{table}%

\subsection{Sky Augmentation and Weather Simulation}

Fig.~\ref{fig:skyrst} shows a group of sky video augmentation results by using our method. Each row shows an input frame from the original video and the processing output at different time steps. The videos in the top-4 rows are downloaded from the YouTube (video source \href{https://www.youtube.com/watch?v=forZrqljb88}{1} and \href{https://www.youtube.com/watch?v=T9mvanUAXV4&t=1152s}{2}) and the video in the last row is captured by ourselves with a handheld smartphone (Xiaomi Mi 8) in Ann Arbor, MI. We generate dynamic sky effects of ``sunset'', ``District-9 ship'', ``super moon'', and ``Galaxy'' for the above video clips. Our method generates visually striking results with a high degree of realism. For more examples and animated demonstrations, please check out our \href{https://jiupinjia.github.io/skyar/}{project website}.

\begin{figure}
    \centering{\includegraphics[width=\linewidth]{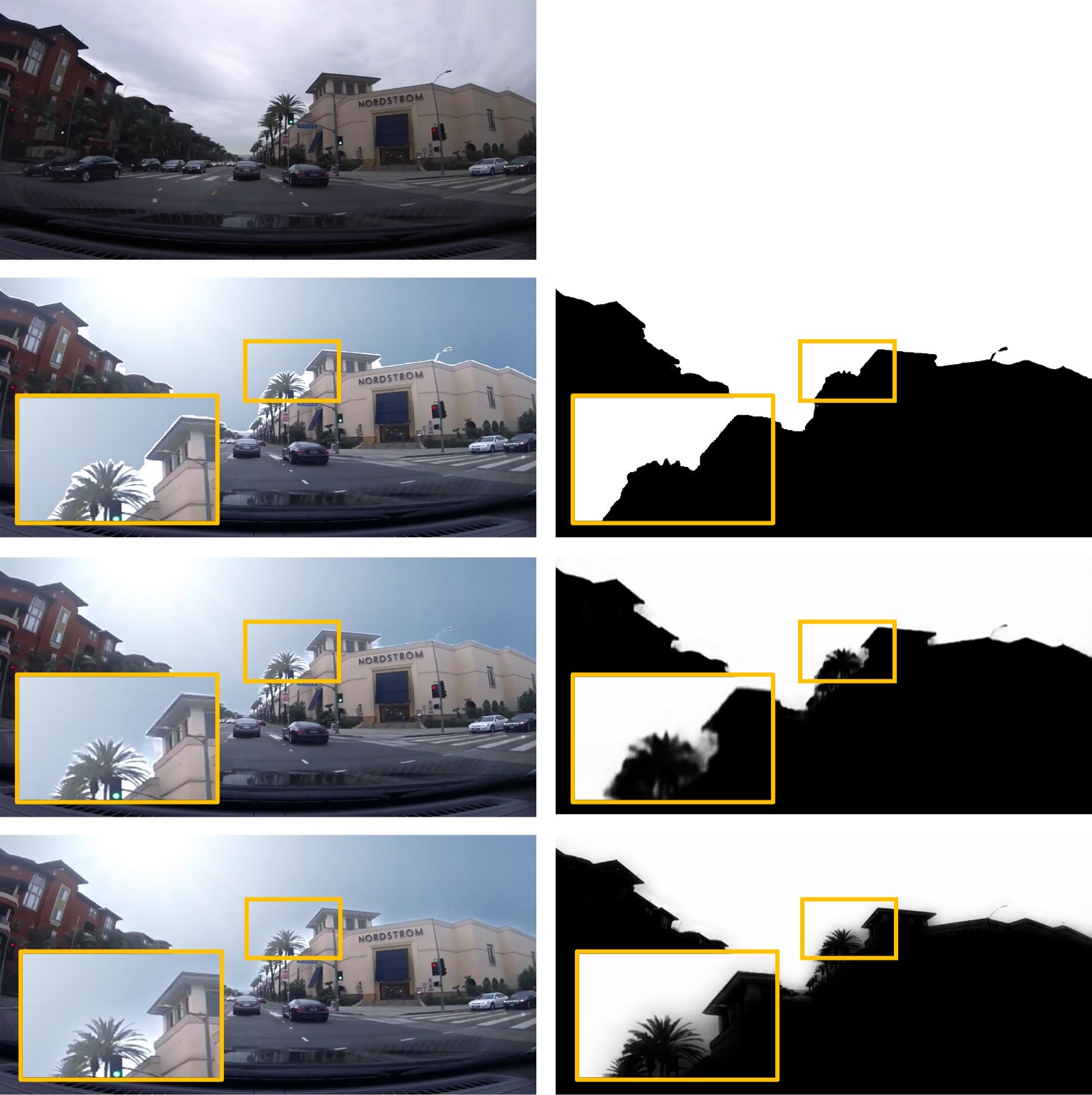}} \\
    \caption{First row: an input frame. Second row: blending result by using hard pixel sky masks. Third and fourth row: blending result based on soft sky matting w/ and w/o applying refinement. We show that the refinement brings much fewer artifacts than the blending results based on coarse sky matte (previously used by \cite{tsai2016sky,mihail2016sky}).}
    \label{fig:mattingcompare}
\end{figure}

Fig.~\ref{fig:weatherrst} shows several groups of results by using our method for weather translation (sunny to cloudy, sunny to rainy, cloudy to sunny, cloudy to rainy). When we synthesize rainy images, we also add a dynamic rain layer (\href{https://www.youtube.com/watch?v=CpS5Ex1Wx-4&t=427s}{video source}) and a haze layer on top of the result by using screen blending~\cite{wikipedia_2020}. We show with only minor modification on the skybox template and the relighting factor, visually realistic weather translation can be achieved. We also compare our method with CycleGAN~\cite{CycleGAN2017}, a well-known unpaired image to image translation method, which is build based on conditional generative adversarial networks. We train CycleGAN on the BDD100K dataset~\cite{bdd100k} which contains outdoor traffic scenes under different weather conditions. Fig.~\ref{fig:cyclegancompare} shows the qualitative comparison results. In Table~\ref{tab:cyclegancompare}, we give a quantitative comparison of the image fidelity of CycleGAN and our method under different weather translation scenarios. Note that since the CycleGAN does not consider temporal dynamics, here we only test on static images. We thus randomly select 100 images from the validation set of BDD100K with corresponding weather conditions for each evaluation group. We introduce two metrics, Perception Index (PI) \cite{blau20182018} and Naturalness Image Quality Evaluator (NIQE)~\cite{mittal2012making}. The two metrics were originally introduced as a no-reference image quality assessment method based on the natural image statistics and are recently widely used for evaluating image synthesis results. We see our method outperforms CycleGAN with a large margin in both quantitative metrics and visual quality.

Another potential application of our method is data augmentation. Domain gap between datasets with limited samples and the complex real-world poses great challenges for data-driven computer vision methods~\cite{patel2015visual}. Even in modern large scale datasets like MS-COCO~\cite{lin2014microsoft} and ImageNet~\cite{deng2009imagenet}, the sampling bias is still severe, limiting the generalization of the model to the real world. For example, domain sensitive visual perceptron models in self-driving may face problems at night or rainy days due to the limited examples in training data. We believe our method has great potential for improving the generalization ability of deep learning models in a variety of computer vision tasks such as detection, segmentation, tracking, etc.

\begin{table}\small
\begin{center}
\begin{tabular}{c|c|ccc}
\toprule
    &  & \multicolumn{3}{c}{Phased time overhead (s)} \\ 
Resolution & Speed & Phase I & Phase II & Phase III \\
\midrule
640$\times$360 pxl & 24.03 fps & 0.0235 & 0.0070 & 0.0111 \\
854$\times$480 pxl & 14.92 fps & 0.0334 & 0.0150 & 0.0186 \\
1280$\times$720 pxl & 7.804 fps & 0.0565 & 0.0329 & 0.0386 \\
\bottomrule
\end{tabular}%
\end{center}
\vspace{-1.5em}
\caption{Speed performance of our method at different output resolutions and the time spent in different processing phases (I. sky matting; II. motion estimation; III. blending).}
\label{tab:speed}%
\end{table}%

\subsection{Speed performance}

Table~\ref{tab:speed} shows the speed performance of our method. The inference speeds are tested on a desktop PC with an NVIDIA Titan XP GPU card and an Intel I7-9700k CPU. The speed at different output resolution and the time spent in different processing stages are recorded. We can see our method reaches a real-time processing speed (24 fps) at the output resolution of 640$\times$320 and a near real-time processing speed (15 fps) at 854$\times$480 but still has large rooms for speed up. As there is a considerable part of the time spent in the sky matting stage, one may easily speed up the processing pipeline by replacing the ResNet-50 with a more efficient CNN backbone, e.g. MobileNet~\cite{howard2017mobilenets,sandler2018mobilenetv2} or EfficientNet~\cite{tan2019efficientnet}.

\begin{table}\small
\begin{center}
\begin{tabular}{l|cc}
\toprule
    & w/ refinement & w/o refinement \\ 
\midrule
w/ positional encoding & 27.31 / 0.924 & 27.17 / 0.929 \\
w/o positional encoding & 27.01 / 0.919 & 26.91 / 0.914 \\
\bottomrule
\end{tabular}%
\end{center}
\vspace{-1.5em}
\caption{Mean pixel accuracy (PSNR / SSIM) of our sky matting model on the dataset~\cite{liba2020sky} w/ and w/o using positional encoding (``CoordConv''s in Table~\ref{tab:encoderdecoder}). The accuracy before and after the refinement is also reported. Higher scores indicate better.}
\label{tab:mattingacc}%
\end{table}%

\subsection{Controlled Experiments}

\textbf{Segmentation vs Soft Matting.} To evaluate the effectiveness of our sky matting network, we design the following controlled experiments where we visually compare the blending results generated by 1) hard pixel segmentation, 2) soft matting before refinement, and 3) soft matting after refinement. We also compare the matting accuracy on the validation set of the dataset~\cite{liba2020sky} (ADE20K+DE+GF). Fig.~\ref{fig:mattingcompare} shows the sky matte and the sky image blending results generated by the above settings. We can see the hard pixel segmentation produces undesired artifact near the sky boundary, and our two-stage matting method can get a more accurate sky region and get a much more visually pleasing blending result than the one stage method. Table~\ref{tab:mattingacc} shows the mean pixel accuracy (PSNR and Structural Similarity (SSIM) index~~\cite{wang2004image}) of our matting model w/ and w/o a second stage refinement. We can see the refinement brings a minor increase in the PSNR but does not necessarily improve the SSIM. Although the improvement is somewhat marginal (PSNR +0.14\%), we find that the visual quality can be greatly improved, especially at the boundary of the sky regions. 

\textbf{Positional embedding.} We also test the matting network without using the coordinate convolution and replace all those ``CoordCoonv'' layers in Table~\ref{tab:encoderdecoder} with conventional convolution layers. We see a noticeable pixel accuracy drop when we remove the coordinate convolution layers (PSNR -0.26 and SSIM -0.015 before refinement; PSNR -0.30 and SSIM -0.005 after refinement), which suggests that the position encoding provides important priors for the sky matting task.

\begin{figure}
    \centering{\includegraphics[width=\linewidth]{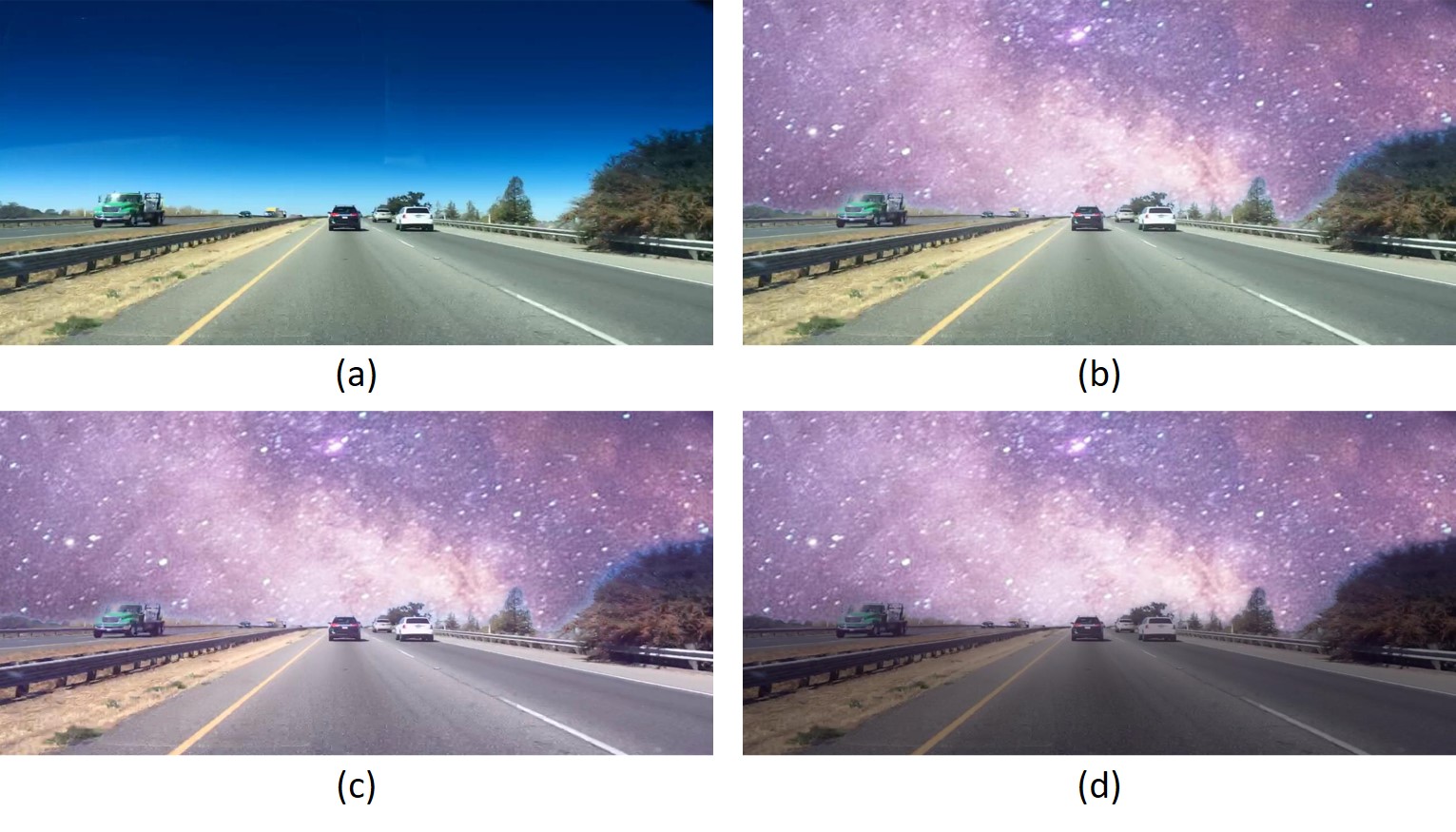}} \\
    \caption{(a) An input frame from BDD100K~\cite{bdd100k}. (b) linear blending result (Eq.~\ref{eq:matting} only). (c) linear blending + color transfer (Eq.~\ref{eq:recoloring:a}). (d) linear blending + color transfer + relighting (Eq.~\ref{eq:recoloring:a} and \ref{eq:recoloring:b}).}
    \label{fig:recoloring}
\end{figure}

\textbf{Color transfer and relighting.} We compare the blending result of our method w/ or w/o using color transfer and relighting. In Fig.~\ref{fig:recoloring}, we give two groups of examples and compare the results of ``linear blending only (Eq.~\ref{eq:matting})'', ``linear blending + color transfer (Eq.~\ref{eq:recoloring:a})'', and `` linear blending + color transfer + relighting (both Eq.~\ref{eq:recoloring:a} and \ref{eq:recoloring:b})''. This results provide an clear demonstration of the significance of our two-step correction - the color transfer can help eliminate the color-tone conflict between foreground and background while the relighting can correct the intensity of ambient light and can further enhances the sense of reality.

\begin{figure}
    \centering{\includegraphics[width=\linewidth]{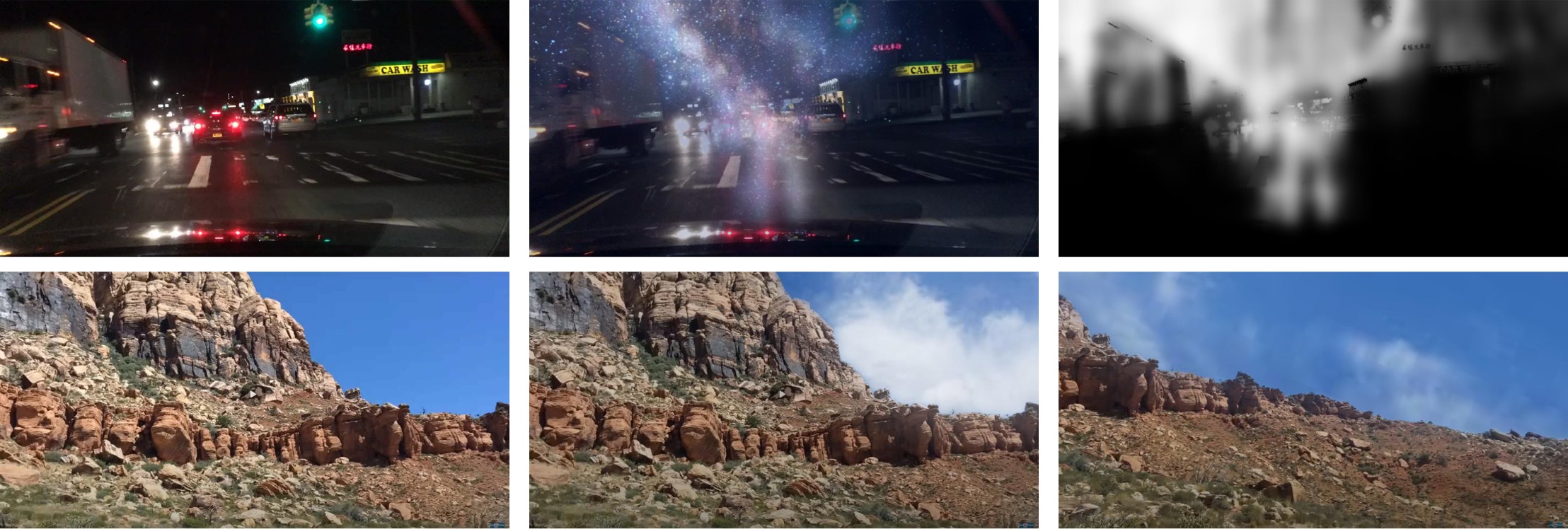}} \\
    \caption{Two failure cases of our method. The top row shows an input frame from BDD100K~\cite{bdd100k} at nighttime (left) and the blending result (middle) produced by wrongly detected sky regions (right). The second row shows an input frame (left, \href{{https://www.youtube.com/watch?v=eK4KqeTFqEA}}{video source}), and the incorrect movement synchronization results between foreground and rendered background (middle and right).}
    \label{fig:falurecases}
\end{figure}

\subsection{Limitation and Future Work.}

The limitation of our method is twofold. First, since our sky matting network is only trained on daytime images, our method may fail to detect the sky regions on nighttime videos. Second, when there are no sky pixels during a certain period of time in a video, or there are no textures in the sky, the motion of the sky background cannot be accurately modeled. This is because the feature points we used for motion estimation are assumed to be located at infinity and using the feature points of objects that are second far away to estimate motion will introduce inevitable errors. Fig.~\ref{fig:falurecases} shows two failure cases of our method. In our future work, we will focus on three directions - the first one is scene-adaptive sky matting, the second one is robust background motion estimation, and the third one is to explore the effectiveness of sky-rendering based data augmentation for object detection and segmentation.

\section{Conclusion}

We investigate sky video augmentation, a new problem in computer vision, namely automatic sky replacement and harmonization in video with purely vision-based approaches. We decompose this problem into three proxy tasks: soft sky matting, motion estimation, and sky blending. Our method does not rely on the inertial measurement unit integrated on the camera devices, and also does not require user interaction. Using our method, users can easily generate highly realistic and cool sky animations in real-time. As a by-product, our method can be also used for image weather translation, and hopefully, it can be used as a new data augmentation approach to enhance the generalization ability of deep learning models in computer vision tasks.

{\small
\bibliographystyle{ieee_fullname}
\bibliography{egbib}
}

\end{document}